\documentclass[10pt,twocolumn,letterpaper]{article}

\usepackage[accsupp]{axessibility}
\usepackage{cvpr}
\usepackage{times}
\usepackage{epsfig}
\usepackage{graphicx}
\usepackage{amsmath}
\usepackage{amssymb}

\usepackage[table,xcdraw]{xcolor}

\begin{document}

\title{Template-based Multi-Domain Face Recognition}

\author{Anirudh Nanduri\\
University of Maryland\\
College Park, MD\\
{\tt\small snanduri@umd.edu}
\and
Rama Chellappa\\
Johns Hopkins University\\
Baltimore, MD\\
{\tt\small rchella4@jhu.edu}
}

\maketitle
\thispagestyle{empty}

\begin{abstract}
Despite the remarkable performance of deep neural networks for face detection and recognition tasks in the visible spectrum, their performance on more challenging non-visible domains is comparatively still lacking. While significant research has been done in the fields of domain adaptation and domain generalization, in this paper we tackle scenarios in which these methods have limited applicability owing to the lack of training data from target domains. We focus on the problem of single-source (visible) and multi-target (SWIR, long-range/remote, surveillance, and body-worn) face recognition task. We show through experiments that a good template generation algorithm becomes crucial as the complexity of the target domain increases. In this context, we introduce a template generation algorithm called Norm Pooling (and a variant known as Sparse Pooling) and show that it outperforms average pooling across different domains and networks, on the IARPA JANUS Benchmark Multi-domain Face (IJB-MDF) dataset. 
\end{abstract}

\section{Introduction}

Owing to the availability of large labeled datasets and computing resources, deep learning algorithms have become the state-of-the-art in many fields. One such is the field of face recognition. The term face recognition in this context broadly covers the following problems: 
1. Face detection: given an image, detect all the faces in it, often the first step in face recognition tasks. 
2. Face verification (1:1 matching): given a pair of faces, verify whether they belong to the same person. An example scenario could be when the live photograph of a person needs to be matched against the photo on their records.
3. Face identification/search (1:N matching): given a face image (probe), recognize the identity of the person by matching the face against a gallery database of $N$ faces. This problem can be further divided into open-set and closed-set protocols. The open-set identification protocol allows for the probes to have identities that may not be present in the gallery, while closed-set protocol assumes all the probe images have a match in the gallery. An example of open-set identification occurs when a photo taken by a CCTV or traffic camera requires the identification of an individual. This photo can be matched against a database of driver's license photos, which may or may not include the individual in question. \par
Face recognition becomes challenging when the images are captured in non-ideal imaging conditions (like rain/snow or low light) or in non-cooperative acquisition conditions (like occlusions, long-distance or extreme poses). Face recognition of cross-spectral images is further challenging because images captured from different spectral bands have different photometric properties. Some of the common non-visual domains used in face recognition \cite{bourlai2012multi} are: near-infrared/NIR (750 \textit{nm} - 1100 \textit{nm}), short-wave infrared/SWIR (1100 \textit{nm} - 2500 \textit{nm}), medium-wave infrared/MWIR (3000 \textit{nm} - 5000 \textit{nm}), and long-wave infrared/LWIR (7000 \textit{nm} - 14000 \textit{nm}). NIR and SWIR fall into the active IR band, meaning that the sensor needs to emit IR light onto the target to capture an image and, MWIR and LWIR comprise the passive or thermal IR band (the sensor primarily detects the emitted radiation). Under low-light conditions, NIR and SWIR can produce images with high SNR when compared to visual (VIS) images. They can also penetrate low levels of atmospheric conditions like rain, fog or smoke. NIR and SWIR are also less susceptible to atmospheric temperature variations or heat hazes compared to thermal images. But MWIR and LWIR images can capture temperature changes from long distances even in the presence of heavy fog or smoke. 


One of the main impediments for multi-domain or cross-spectral face recognition is the lack of large labeled datasets for domains such as long-range visible, visible surveillance, NIR, SWIR, MWIR and LWIR. Face detection/recognition on visible images has improved significantly over the years because of the large publicly available datasets such as WebFace260M \cite{zhu2022webface260m}, Glint360k \cite{an2022pfc}, CelebA \cite{liu2015faceattributes}, MS-Celeb-1M \cite{guo2016ms}, UMD Faces \cite{bansal2017umdfaces}, CASIA-WebFace \cite{yi2014learning}, VGGFace \cite{parkhi2015deep}, \cite{cao2018vggface2}, and WIDER Face \cite{yang2016wider}. Some of the popular IR face datasets include ARL-VTF \cite{poster2021large}, IJB-MDF \cite{kalka2019iarpa}, Equinox \cite{socolinsky2001illumination}, NVIE \cite{wang2010natural}, CASIA NIR-VIS 2.0 \cite{li2013casia}, LDHF-DB \cite{maeng2012nighttime}, \cite{mallat2018benchmark} and \cite{hu2016polarimetric}. Table \ref{tab:datasets} compares the sizes of the VIS datasets and IR datasets. 

Domain adaptation and generalization methods require a (large) training dataset to be effective. In most cases, they also require paired source-target data - that is, they require (or greatly benefit from) having the same identities in both the source and target domains. But these requirements are very difficult to satisfy in many real world scenarios. In this paper, we investigate how the domain shift can be somewhat offset by using pre-trained models trained on very wide VIS datasets (with a lot of identities) and through a novel template generation algorithm, when there is \textit{no access to training data}.

Subject-specific modeling or template-based face recognition was first introduced by the IJB-A dataset \cite{klare2015pushing}. In this paradigm, all the images and/or videos corresponding to a subject are combined into a single template for matching. Compared to pair-wise comparison methods \cite{taigman2014deepface}, \cite{schroff2015facenet}, template-based approaches are computation and memory efficient. An obvious choice for generating the templates is to take the average of all the feature representations of the media. This is termed average pooling \cite{chen2018unconstrained}, \cite{ranjan2017all}, \cite{sankaranarayanan2016triplet}, \cite{parkhi2015deep} and its main drawback is that it gives equal weight to all pieces of media, irrespective of their quality. We observe that typically, templates from challenging domains have more variation in face quality and more low-quality faces. As such, a good template generation algorithm becomes crucial for face recognition in these domains.  

\begin{table}[t]
	\renewcommand{\arraystretch}{0.8}
	\caption{\label{tab:datasets}Comparison of the Sizes of Different Face Datasets in the VIS and IR Domains.}
	\centering
		\resizebox{\columnwidth}{!}{
		\renewcommand{\arraystretch}{1.2}
	\begin{tabular}{c c c} 
		\hline
		\rowcolor[rgb]{0.753,0.753,0.753} \textbf{Dataset} & \textbf{~ Domain(s)} & \textbf{Number of subjects}  \\ 
		\hline
		WebFace260M       & VIS 				  & 4,000,000  \\
		Glint360k         & VIS					  & 360,000    \\
		MS-Celeb-1M       & VIS                   & 100,000    \\
		CASIA WebFace     & VIS                   & 10,575     \\
		CelebA            & VIS                   & 10,177     \\
		UMD Faces         & VIS                   & 8,277      \\
		VGG Face          & VIS                   & 2,622      \\
		\hline
		CASIA NIR-VIS 2.0 & VIS, NIR              & 725        \\
		ARL-VTF           & VIS, LWIR             & 395        \\
		IJB-MDF           & VIS, SWIR, MWIR, LWIR, Long-range & 251        \\
		NVIE              & VIS, LWIR             & 103        \\
		LDHF-DB           & VIS, NIR              & 100        \\
		Equinox           & VIS, LWIR             & 91         \\ \hline
	\end{tabular}}
	
\end{table}

The IJB-MDF dataset \cite{kalka2019iarpa} comprises of images and videos captured using a variety of cameras: fixed and body-worn, capable of imaging at visible, short-wave, mid-wave and long-wave infrared wavelengths at distances up to 500m. It includes face detection, 1:N closed-set and open-set identification protocols. Two disjoint galleries (visible domain) are provided to evaluate the 1:N identification protocols with probes from SWIR, MWIR, LWIR, visible-remote and visible-body-worn (GoPro) domains.

The following are the main contributions of this paper:
\begin{itemize}
	\item We introduce a template generation algorithm called Norm Pooling, along with some of its variants.
	\item We enhance the benchmark performance on the IJB-MDF dataset and also propose a new 1:N identification protocol with larger template sizes.
	\item We conduct extensive experiments with three different pretrained models (trained with different loss functions) and show that the proposed template generation algorithm outperforms the standard average pooling.
\end{itemize}

The rest of the paper is organized as follows: Section 2 details some works related to cross-spectral face recognition and template generation; Section 3 describes the face recognition pipeline along with template generation algorithms; in Section 4 we present experiments and results; and finally Section 5 contains conclusions. 

\section{Related Work}


\subsection{Cross-spectral Face Recognition}

Bourlai et al. \cite{bourlai2010cross} published one of the first papers which looked into the problem of cross-spectral SWIR face recognition. They collected the WVU Multispectral dataset and presented cross-spectral matching results using classical face recognition methods like PCA with k-NN. Kalka et al. \cite{kalka2011cross} 
extended the work in \cite{bourlai2010cross} to heterogenous face recognition in semi-controlled and uncontrolled environments. 
Bourlai et al. \cite{bourlai2012multi} 
studied SWIR-VIS, MWIR-MWIR, MWIR-VIS and NIR-VIS matching and extended the work presented in \cite{kalka2011cross} to more challenging scenarios (cross-distance matching) and other domains like MWIR and NIR. Maeng et al. \cite{maeng2012nighttime} 
collected the Long Distance Heterogeneous Face Database, proposed Gauss-SIFT algorithm and reported results on both intra-spectral and cross-spectral cross-distance matching. Juefei-Xu et al. \cite{juefei2015nir} 
proposed a dictionary learning approach to learn a mapping function between VIS and NIR domains, thus reducing the problem of cross-spectral matching to intra-spectral matching. Lezama et al. \cite{lezama2017not} proposed a deep learning-based approach by adapting a deep network pre-trained on VIS images to generate discriminative features from both  VIS and NIR images. 
Song et al. \cite{song2018adversarial} proposed a deep network with cross-spectral face hallucination and discriminative feature learning for VIS-NIR matching using a GAN, by employing an adversarial loss and a high-order variance discrepancy loss to measure the global and local discrepancy between the domains. He et al. \cite{he2019adversarial} 
extended the work reported in \cite{song2018adversarial} by performing cross-spectral face hallucination using inpainting of VIS image textures from NIR textures and, pose correction to generate VIS images at frontal pose. Fu et al. \cite{fu2021dvg} proposed a Dual Variational Generation framework to learn the joint distribution of paired heterogeneous images, and then generated paired images from the two domains. These generated images are used to train a face recognition network using a contrastive learning mechanism. Peri et al. \cite{peri2021synthesis} proposed a synthesis-based approach using GAN architectures for thermal-to-visible face verification. 

\subsection{Template Generation Approaches}
Parkhi et al. \cite{parkhi2015deep} tackled video-based face recognition on YouTube Faces (YTF) dataset \cite{wolf2011face} by computing the average of the top K face features ranked by their facial landmark confidence score in each video. Crosswhite et al. \cite{crosswhite2018template} studied the problem of template adaptation on the IJB-A dataset \cite{klare2015pushing} using a combination of deep networks and linear SVMs. Using features from deep networks, they trained one-vs-rest linear SVM classifiers using all the media in a template as positive features to classify the new subjects. To construct the template feature, they unit normalized the mean of media features. Yang et al. \cite{yang2017neural} proposed Neural Aggregation Network (NAN) for the purpose of generating the template feature from a face video or an image set. Their aggregation module used two cascaded attention blocks that can take any number of feature vectors as input and produces a single template feature with the same dimensions. Ranjan et al. \cite{ranjan2018crystal} proposed Quality Pooling that uses face detection probability scores to calculate the weights for each media in the template. An obvious limitation of deep network-based methods like NAN is the need for additional training, which requires labeled training data. Jawade et al. \cite{jawade2023conan} proposed CoNAN which uses a learned face context vector that is conditioned over the distribution of the media features to compute the weights.

In contrast to most of these previous works which use some metadata information or train an aggregation network to generate the template feature, we propose a simpler approach that can be applied even in the absence of training data or metadata. The protocols we consider in this work do not provide any training data. \textit{Therefore, to generate the templates, we are limited to algorithms that perform operations exclusively on media features without any additional training}.

\section{Face Recognition Pipeline}

\begin{figure*}[h]
	\centering
	\includegraphics[width=0.9\textwidth]{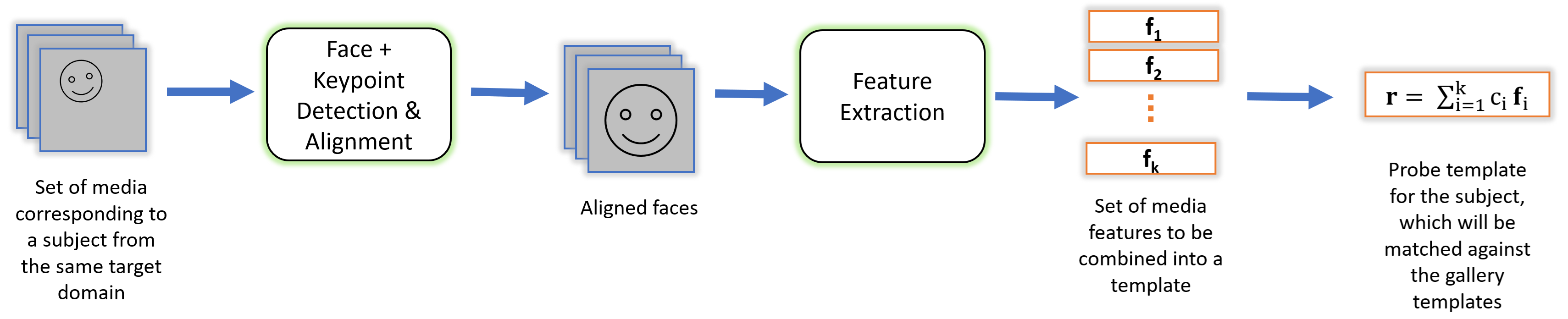}
	\caption{A typical pipeline for generating a face template from a set of media (images/frames) corresponding to a subject. The template generation algorithm determines the coefficient $c_i$ for each media feature $\textbf{f}_i$. The resulting probe template corresponding to a subject is matched against the gallery templates to determine the identity of the subject. We generate one template per subject per domain.} 
	\label{fig:pipeline}
	\centering
\end{figure*}

A typical (image-based) face recognition pipeline \cite{ranjan2019fast} consists of face detection, landmark localization and face alignment, feature extraction and feature matching. An input image is first passed through the face detector to predict a bounding box around the detected face. Then facial key-points are computed in the detected face and are used to align the faces to make them invariant to rotation and scaling. The feature extractor generates feature vectors from the aligned faces and finally, to identify a face, its features are matched against the features of gallery faces.\par

Video data is usually processed by extracting the individual frames and passing them through a face detector \cite{zheng2020automatic}. Deep features of the detected faces are then extracted by passing them through the pipeline described above. Face templates with unique identities are then constructed by using either face association (for multi-shot videos) or face tracking (for single-shot videos) algorithms.

In this paper, we follow the subject-specific modeling paradigm mentioned in \cite{kalka2019iarpa}. We generate one probe template for each domain, for each subject, using the template ids provided in the protocol. This alleviates the need for a face association algorithm and simplifies the problem formulation. Fig. \ref{fig:pipeline} illustrates this pipeline.

We now describe the various modules used in our face recognition pipeline.

\subsection{Face Detection}
Face detection is the first module in any face recognition pipeline. We employ the SCR (sample and computation redistribution) face detector presented in \cite{guo2021sample}. Sample redistribution refers to augmenting training samples for shallow stages and computation redistribution is a two-step method to obtain an optimized network design across the network's backbone, neck and head. The computation redistribution involves network structure search: the backbone is based on RetinaNet \cite{lin2017feature}, the neck is based on a Path Aggregation Feature Pyramid Network (PAFPN) \cite{liu2018path} and the head is a stack of $3 \times 3$ convolutional layers. 

\begin{figure*}[h]
	\centering
	\includegraphics[width=0.8\textwidth]{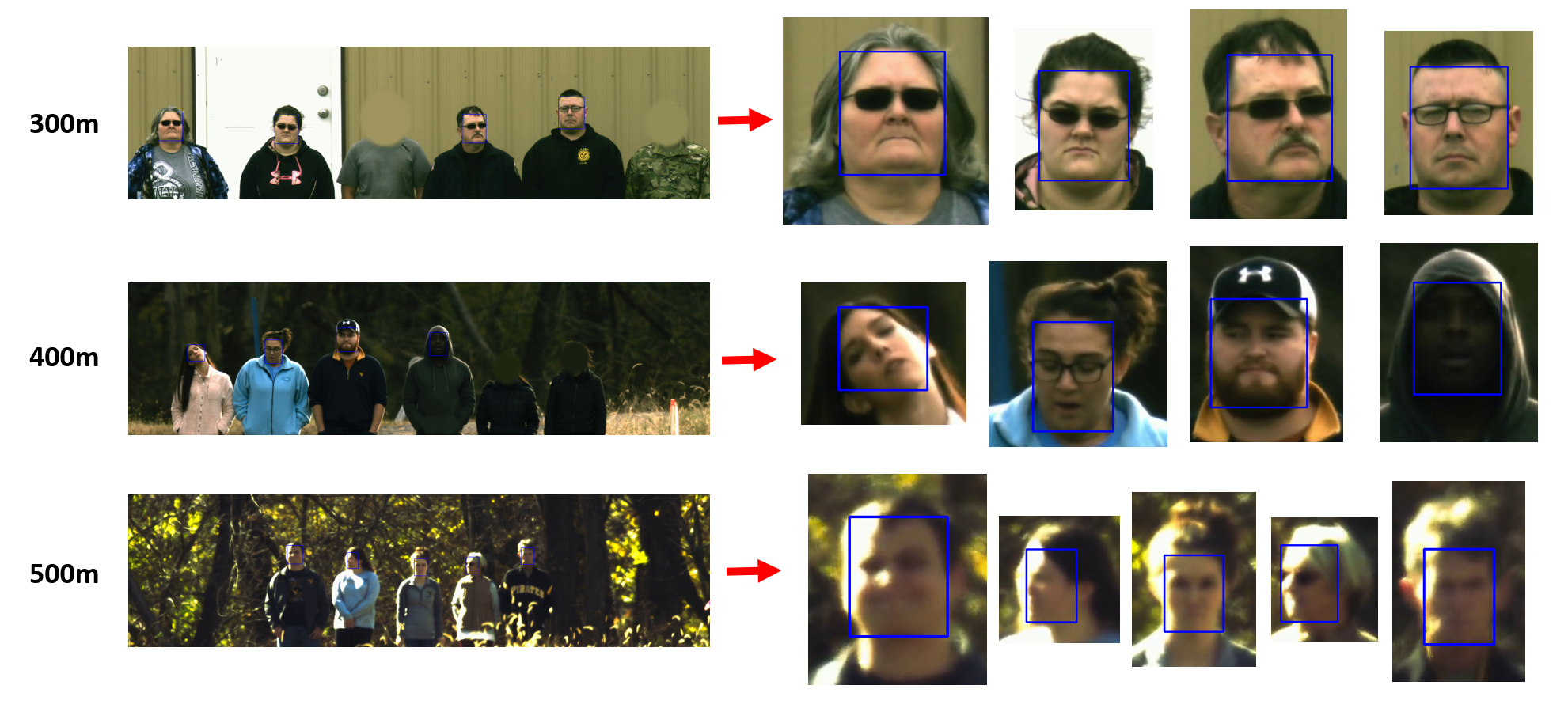}
	\caption{Face detection results on visible images captured at 300, 400, and 500m in the IJB-MDF dataset. Left: Original images. Right: Face detection results.} 
	\label{fig:detection_remote_vis}
	\centering
\end{figure*}

Fig. \ref{fig:detection_remote_vis} shows some example detection results on visible images captured at distances of 300, 400 and 500m from the IJB-MDF dataset. In most of our experiments with the 1:N identification protocol, we use the ground-truth face detections to enable a fair comparison with previous works \cite{kalka2019iarpa} that reported results on the IJB-MDF dataset. 

\subsection{Keypoint Detection and Alignment}

Face keypoints include centers and corners of eyebrows, eyes, nose, mouth, earlobes and chin. Adaptive Wing Loss \cite{wang2019adaptive} based model (AWing) is used for keypoint detection. Adaptive Wing loss can adapt its curvature based on whether the ground truth heatmap pixels belong to the foreground or background, so that there is more focus on the foreground and difficult background pixels, and less focus on the background pixels.


The detected keypoints are then used to transform the face image, ensuring that the eyes, nose and mouth are mapped to standardized locations. This alignment step introduces a degree of pose invariance, improving the overall performance of the face recognition system.

\subsection{Feature Extraction}

For the feature extraction module, we use Resnet-100 models trained using CosFace \cite{wang2018cosface}, ArcFace \cite{deng2019arcface}, and AdaFace \cite{kim2022adaface} losses. 

AdaFace loss is an adaptive margin loss that emphasizes hard samples if the image quality is high and ignores very hard samples if the image quality is low. Image quality is estimated using the normalized feature norm as defined in (\ref{eq:ada1}).

\begin{equation} \label{eq:ada1}
	\widehat{\|\textbf{z}_i\|} = \left \lfloor \frac{\|\textbf{z}_i\| - \mu_z}{\sigma_z / h} \right \rceil _{-1} ^{1},
\end{equation}
where $\mu_z$ and $\sigma_z$ are the mean and standard deviation of all the feature norms $\|\textbf{z}_i\|$ in a batch, $h$ is a hyperparameter, $\lfloor . \rceil$ clips the values to be in the range $[-1, 1]$.

The AdaFace loss is then given as in (\ref{eq:ada2}) and (\ref{eq:ada2_2}). 

\begin{equation} \label{eq:ada2}
	L_{AdaFace} = - \log \frac{e^{f(\theta_{y_i},m)}}{e^{f(\theta_{y_i},m)} + \sum_{j\ne y_i}^n e^{s \cos \theta_j}},
\end{equation}
where $f(.)$ is the margin function which is defined as in (\ref{eq:ada2_2}), $y_i$ is the index of the ground-truth label, $\theta_j$ is the angle between the feature vector $\textbf{z}_i$ and the $j^{th}$ classifier weight vector $\textbf{W}_j$, $s$ is the scaling factor for the feature vector $\textbf{z}_i$ after $l_2$ normalization, and $m$ is the margin.

\begin{equation} \label{eq:ada2_2}
	f(\theta_j,m) = \begin{cases}
		s \cos (\theta_j + g_{\text{angle}})-g_{\text{add}} & j = y_i\\
		s \cos \theta_j & j \ne y_i
	\end{cases}       ,
\end{equation}
where $g_{\text{angle}}$ and $g_{\text{add}}$ are defined as: $g_{\text{angle}} = -m.\widehat{\|\textbf{z}_i\|}$ and $g_{\text{add}} = m.\widehat{\|\textbf{z}_i\|} + m$.

CosFace loss and ArcFace losses can be defined as special cases of the AdaFace loss when $\widehat{\|\textbf{z}_i\|} = 0$, and $\widehat{\|\textbf{z}_i\|} = -1$ respectively.

\subsection{Template Generation}

In the subject-specific modeling paradigm, the set of all media (images and/or videos) corresponding to a subject are combined into a single template for matching. We refer to this as template generation or template pooling.

Given a set of features ${\textbf{f}_1, \textbf{f}_2, ...\textbf{f}_k}$ corresponding to face detections in frames/images of a subject, the feature vector $\textbf{r}$ for the resultant template can be defined as in (\ref{eq:t1})

\begin{equation} \label{eq:t1}
	\textbf{r} = \sum_{i=1}^{k} c_i\textbf{f}_i ,
\end{equation}
where $c_i$ corresponds to the weight given to the $i^{th}$ frame/image in the final template. 

\textbf{Average Pooling (AP):}
When $c_i = 1/k$, the template generation algorithm is termed Average Pooling. This is the simplest and most commonly used method for generating templates \cite{kalka2019iarpa}. 

\textbf{Norm Pooling (NP):}
Norm pooling computes the weights $c_i$ as in (\ref{eq:t8}) and (\ref{eq:t4}), where $\lambda$ is a hyperparameter. The intuition behind this formulation is that $l_i$ approximates the image quality using the feature norm  (\cite{parde2016deep}, \cite{ranjan2018crystal}, \cite{kim2022adaface}, \cite{terhorst2023qmagface}), thus giving more weight to higher quality media in the template. 

\begin{equation} \label{eq:t8}
	c_i = \mathrm{softmax}(\lambda l_i),
\end{equation}

\begin{equation} \label{eq:t4}
	l_i = \mathrm{max\text{-}normalized}(\|\textbf{f}_i\|) = \frac{\|\textbf{f}_i\|}{\max\limits_j \|\textbf{f}_j\|}
\end{equation}

Kim et al. \cite{kim2022adaface} showed that feature norm can be used as a proxy for image quality (especially in models trained with a margin-based softmax loss) for faces in the visible domain by computing the Pearson correlation coefficients between the BRISQUE \cite{mittal2012no} image quality scores and the feature norms. Table \ref{tab:brisque} presents the results of similar experiments conducted across various models and domains like long-range visible and SWIR (from the IJB-MDF dataset). Although the correlation scores indicate only a weak to moderate correlation, the reason could be that BRISQUE is designed mainly to detect the 'naturalness' in images due to the presence of distortions like white noise and Gaussian blur.

\begin{table}[t!]
	\renewcommand{\arraystretch}{1.5}
	\caption{\label{tab:brisque}Pearson correlation coefficients [-1, 1] between feature norms and BRISQUE image quality scores across various domains}
	\centering
	\resizebox{1.\columnwidth}{!}{
		\renewcommand{\arraystretch}{1.8}
		\begin{tabular}{c c c c c c} 
			\hline
			\rowcolor[rgb]{0.753,0.753,0.753} \textbf{Model} &  \textbf{VIS-500m} & \textbf{VIS-400m} & \textbf{VIS-300m} & \textbf{SWIR-15m} & \textbf{SWIR-30m}  \\ 
			\hline
			CosFace  & 0.42 & 0.42 & 0.22 & 0.23 & 0.27 \\
			AdaFace & 0.33 & 0.24 & 0.27 & 0.24 & 0.31 \\
			ArcFace & 0.41 & 0.38 & 0.23 & 0.21 & 0.26 \\

			\hline
	\end{tabular}}
\end{table}

\textbf{Norm Pooling variant (NP$^*$):}
Min-max normalization of the norm performs better than max-normalization (\ref{eq:t4}) when there is not enough spread in the distribution of norms. We refer to this variant as NP$^*$ and it is defined as in (\ref{eq:t7})
\begin{equation} \label{eq:t7}
	\begin{split}
		l_i^* &= \mathrm{min\text{-}max\text{-}normalized}(\|\textbf{f}_i\|) \\
		&= \frac{\|\textbf{f}_i\| - \min\limits_j \|\textbf{f}_j\|}{\max\limits_j \|\textbf{f}_j\| - \min\limits_j \|\textbf{f}_j\|}
	\end{split}
\end{equation}

\textbf{Sparse Pooling (SP):}
We propose another variant of norm-pooling called sparse-pooling, where sparsemax \cite{martins2016softmax} function is used instead of the softmax function to calculate the weights $c_i$ as in (\ref{eq:t9}). 
\begin{equation} \label{eq:t9}
	c_i = \mathrm{sparsemax}(\lambda l_i)
\end{equation}

The sparsemax \cite{martins2016softmax} function has the ability to produce sparse distributions where some of the output probabilities can be exactly zero, unlike the softmax function. This lets us discard poor quality media when creating the template. Sparsemax is defined as in (\ref{eq:t5})

\begin{equation} \label{eq:t5}
	\mathrm{sparsemax}(\mathbf{z}) = \underset{\mathbf{p} \in \Delta^{k-1}}{\text{arg min}} \|\mathbf{p} - \mathbf{z}\|^2,
\end{equation}
where \(\mathbf{z}\) is the input vector to the sparsemax function, \(\mathbf{p}\) represents a probability distribution (a point in the probability simplex \(\Delta^{k-1}\)), and \(\|\mathbf{p} - \mathbf{z}\|^2\) is the squared Euclidean distance between \(\mathbf{p}\) and \(\mathbf{z}\). The Sparsemax function finds the probability distribution \(\mathbf{p}\) within the simplex that is closest to the input vector \(\mathbf{z}\) in terms of the Euclidean distance. Sparsemax can result in sparse distributions because \(\mathbf{p}\) can hit the boundary of the simplex. The closed-form solution for sparsemax can be found in \cite{martins2016softmax}.

\section{Experiments and Results}

\textbf{IJB-MDF Dataset:}
The IARPA JANUS Benchmark Multi-domain Face (IJB-MDF) dataset consists of images and videos of 251 subjects captured using a variety of cameras corresponding to visible, short-, mid-, and long-wave infrared and long range surveillance domains. There are 1,757 visible enrollment images,  40,597 short-wave infrared (SWIR) enrollment images and over 800 videos spanning 161 hours.


Following is the list of domains in the IJB-MDF dataset:

\begin{itemize}
	\setlength\itemsep{-0.5em}
	\item Domain 0: Visible enrollment
	\item Domain 1: Visible surveillance
	\item Domain 2: Visible gopro
	\item Domain 3: Visible 500m
	\item Domain 4: Visible 400m
	\item Domain 5: Visible 300m
	\item Domain 6: Visible 500m 400m walking
	\item Domain 7: MWIR 15m
	\item Domain 8: MWIR 30m
	\item Domain 9: LWIR 15m
	\item Domain 10: LWIR 30m
	\item Domain 11: SWIR enrollment nofilter
	\item Domain 12: SWIR enrollment (captured at 1150 nm)  
	\item Domain 13: SWIR enrollment (captured at 1350 nm)
	\item Domain 14: SWIR enrollment (captured at 1550 nm)    
	\item Domain 15: SWIR 15m
	\item Domain 16: SWIR 30m
\end{itemize}

\begin{table*}[t]
	\renewcommand{\arraystretch}{1.2}
	\caption{\label{tab:verification_results}1:N Closed-Set Identification Results with the Exhaustive Protocol - Rank-1 Retrieval Rates (\%)}
	\centering
	\resizebox{2\columnwidth}{!}{
		\renewcommand{\arraystretch}{1.2}
		\begin{tabular}{c c c c c c c c} 
			\hline
			\rowcolor[rgb]{0.753,0.753,0.753} 
			\textbf{Model} & \textbf{VIS-Surv} & \textbf{VIS-GoPro} & \textbf{VIS-500m} & \textbf{VIS-400m} & \textbf{VIS-300m} & \textbf{SWIR-15m} & \textbf{SWIR-30m}  \\ 
			\hline
			CosFace Glint360k (AP) & 74.90 & 66.67 & 81.12 & 88.89 & 99.45 & 90.00 & 66.93 \\
			CosFace Glint360k (NP: $\lambda = 10$) & \textbf{98.01} & \textbf{83.33} & \textbf{86.01} & \textbf{91.92} & \textbf{100.00} & \textbf{91.60} & \textbf{70.52} \\
			\hline
			ArcFace MS1MV3 (AP) & 81.67 & 64.29 & 77.62 & 89.90 & 100.00 & 84.00 & \textbf{66.53} \\
			ArcFace MS1MV3 (NP: $\lambda = 10$) & \textbf{96.41} & \textbf{78.57} & 77.62 & \textbf{91.92} & 100.00 & \textbf{86.00} & 66.14 \\
			\hline
			AdaFace WebFace12M (AP) & 63.75 & 59.52 & 90.21 & 96.46 & 100.00 & 94.40 & 77.29 \\
			

			AdaFace WebFace12M (NP: $\lambda = 50$) & \textbf{81.27} & \textbf{61.90} & \textbf{90.91} & \textbf{97.98} & 100.00 & \textbf{95.60} & \textbf{79.68} \\
			
			\hline
	\end{tabular}}
\end{table*}

\begin{table*}[t]
	\renewcommand{\arraystretch}{1.2}
	\caption{\label{tab:verification_results_legacy}1:N Closed-Set Identification Results with the Legacy Protocol from \cite{kalka2019iarpa} - Rank-1 Retrieval Rates (\%)}
	\centering
	\resizebox{2\columnwidth}{!}{
		\renewcommand{\arraystretch}{1.2}
	\begin{tabular}{c c c c c c c c} 
		\hline
		\rowcolor[rgb]{0.753,0.753,0.753} \textbf{Model} & \textbf{VIS-Surv} & \textbf{VIS-GoPro} & \textbf{VIS-500m} & \textbf{VIS-400m} & \textbf{VIS-300m} & \textbf{SWIR-15m} & \textbf{SWIR-30m}  \\ 
		\hline
		CosFace Glint360k (AP) & 28.30 & 66.67 & \textbf{73.08} & 92.31 & 100.00 & 82.40 & 60.16 \\
		CosFace Glint360k (NP: $\lambda = 10$) & \textbf{40.25} & \textbf{75.00} & 69.23 & 92.31 & 100.00 & \textbf{85.60} & \textbf{63.35} \\ 
		\hline
		ArcFace MS1MV3 (AP) & 28.30 & 75.00 & 69.23 & 84.62 & 100.00 & 76.00 & 56.57 \\
		ArcFace MS1MV3 (NP: $\lambda = 10$) & \textbf{36.48} & 75.00 & \textbf{73.08} & 84.62 & 100.00 & \textbf{77.60} & \textbf{57.37} \\ 
		
		\hline
		AdaFace WebFace12M (AP) & 36.48 & 58.33 & 76.92 & 96.15 & 100.00 & 86.80 & 68.13 \\

		AdaFace WebFace12M (NP: $\lambda = 50$) & \textbf{40.88} & \textbf{66.67} & 76.92 & 96.15 & 100.00 & \textbf{87.20} & \textbf{70.12} \\ 
		\hline
		Baseline \cite{kalka2019iarpa} & 20.12 & 62.85 & 66.25 & 77.50 & 100.00 & 72.28 & 56.11 \\
		\hline
	\end{tabular}}
\end{table*}

\if false
Specifically, we use the following three models:

\begin{itemize}
	\item Resnet-100 network trained on Glint360k with CosFace loss
	\item Resnet-100 network trained on MS1MV3 with ArcFace loss
	\item Resnet-100 network trained on WebFace-12M with AdaFace loss
\end{itemize}
\fi

The IJB-MDF dataset provides three different ground-truth files that are referred to as - the face detection legacy protocol, 1:N identification legacy protocol and the end-to-end protocol. The baseline results provided in the dataset release paper \cite{kalka2019iarpa} use the legacy protocol. The legacy protocol is constructed using a very small subset of frames from all the videos (across all domains) and has only about 32k detections in the face detection protocol and about 51k labeled faces in the 1:N identification protocol. In contrast, the ground-truth file for the end-to-end protocol has about 2.3M labeled faces along with their bounding boxes. Although the name implies that this ground-truth was supposed to be used for evaluating the end-to-end performance of the face recognition system, we limit the scope of this paper to evaluating face detection and identification separately. So we propose a new 'exhaustive protocol' for 1:N identification that uses all the detected faces from the end-to-end protocol to generate the templates. 

In most of our experiments, we use state-of-the-art Resnet-100 models trained on Glint360k \cite{an2022pfc}, MS1MV3 \cite{guo2016ms, deng2019arcface, deng2019lightweight} and WebFace \cite{zhu2022webface260m} datasets with CosFace, ArcFace and AdaFace loss functions respectively, as our feature extractors.

\subsection{1:N Face Identification}	
Since the IJB-MDF dataset has data from very challenging domains for face detectors, for the sake of consistency with the reported results in \cite{kalka2019iarpa}, we use the ground-truth detection bounding boxes to obtain cropped faces. They are then passed through the AWing model to obtain the keypoints, which are then used to perform face alignment. We then pass the aligned faces through our feature extraction module to obtain 512-dimensional features. Feature templates are generated for each subject (one per domain) using the template pooling algorithm. 
For the 1:N identification protocols, there are two disjoint galleries G1 and G2 with 126 and 125 subjects respectively, to facilitate both closed-set (searches where all the probe templates have a corresponding mate in the gallery) and open-set (where a probe template need not have a corresponding mate in the gallery) identification scenarios. The gallery images correspond to the visible enrollment domain. The probe domains considered in this paper are: visible surveillance, visible go-pro, visible 500m, visible 400m, visible 300m, SWIR at 15m and SWIR at 30m. We exclude the MWIR and LWIR domains in our experiments because the domain gap is too large and the networks trained on visible domain images are unable to extract meaningful features on these domains.

\subsubsection{Legacy Protocol}
The evaluation protocol in \cite{kalka2019iarpa} uses a very small subset of frames from each video for generating the probe templates. They randomly select between 1 and 30 faces for each subject, for each domain. We report the rank-1 retrieval rates (obtained using the three models described above) with this protocol in Table ~\ref{tab:verification_results_legacy}, with both average pooling (AP) and norm pooling (NP). We also include the best baseline reported in \cite{kalka2019iarpa}, which uses average pooling. From the results we notice that norm pooling outperforms or matches average pooling with all the models on all the domains (except the CosFace model on VIS-500m). Our results also show significant improvement over the baseline. Among the three models, AdaFace-WebFace12M performed the best on all the domains, except the VIS-GoPro domain. We hypothesize that this is because of two reasons - (i) the AdaFace model is trained to perform well on low-quality and small faces, and except for the GoPro domain (which has images captured at a distance of 2-5m), all the other domains have comparatively poor quality faces, and (ii) the training dataset for the AdaFace model is much larger than that of the CosFace and ArcFace models. This suggests that pretraining on a larger, wider dataset helps bridge the domain gap, especially when the template size is small.

\begin{table*}[t]
	\renewcommand{\arraystretch}{1.2}
	\caption{\label{tab:e2e_results_t}Identification with Detected Faces using Average Pooling (AP), Quality Pooling (QP), and Norm Pooling (NP) - Rank-1 Retrieval Rates (\%)}
	\centering
	\resizebox{2\columnwidth}{!}{
		\renewcommand{\arraystretch}{1.2}
		\begin{tabular}{c c c c c c c c} 
			\hline
			\rowcolor[rgb]{0.753,0.753,0.753} \textbf{Model} & \textbf{VIS-Surv} & \textbf{VIS-GoPro} & \textbf{VIS-500m} & \textbf{VIS-400m} & \textbf{VIS-300m} & \textbf{SWIR-15m} & \textbf{SWIR-30m}  \\ 
			\hline
			CosFace Glint360k + AP & 99.60 & 94.12 & 82.52 & 90.91 & 100.00 & 93.98 & 76.10 \\
			CosFace Glint360k + QP & 99.60 & 94.12 & 83.22 & 91.41 & 100.00 & 93.98 & 76.10 \\
			CosFace Glint360k + NP & \textbf{100.00} & \textbf{97.06} & \textbf{86.71} & \textbf{93.94} & 100.00 & \textbf{95.18} & \textbf{78.49} \\
			\hline
			ArcFace MS1MV3 + AP & 99.20 & 94.12 & 78.32 & 92.42 & 100.00 & 90.76 & 71.31 \\
			ArcFace MS1MV3 + QP & 99.20 & 94.12 & 78.32 & 92.42 & 100.00 & 90.76 & 71.31 \\
			ArcFace MS1MV3 + NP & \textbf{99.60} & 94.12 & \textbf{79.02} & \textbf{92.93} & 100.00 & 90.76 & \textbf{72.91} \\
			\hline
			AdaFace WebFace12M + AP & 99.60 & 91.18 & 92.31 & 97.98 & 99.45 & 97.19 & 85.66 \\
			AdaFace WebFace12M + QP & 99.60 & 91.18 & 92.31 & 97.98 & 99.45 & 97.19 & 85.66 \\
			AdaFace WebFace12M + NP & \textbf{100.00} & 91.18 & \textbf{93.01} & \textbf{98.99} & \textbf{100.00} & \textbf{97.99} & \textbf{88.84} \\
			\hline
	\end{tabular}}
\end{table*}


\subsubsection{Exhaustive Protocol}
We also conduct experiments using all the detected faces of a subject (for each domain separately) to generate the probe template. In this protocol, the template sizes (number of media in a template) are much larger compared to the legacy protocol - on average, about 45 times more. Table ~\ref{tab:verification_results} shows the rank-1 retrieval rates with this protocol. Again, norm pooling outperforms or matches the performance of average pooling across all the domains with all three networks (except for the ArcFace model on SWIR-30m). The exception can be explained by the fact that the optimal value of $\lambda$ that maximizes performance depends on both the network as well as the domain. With a $\lambda$ value of 5 instead of 10, the ArcFace model has a rank-1 retrieval rate of 67.73 (with NP), which is higher than the retrieval rate with average pooling (66.53). 
The model trained with CosFace loss on Glint360k dataset has the best performance on the VIS-Surv and VIS-GoPro domains, while the AdaFace model trained on WebFace12M performs the best on remote domains - visible-500, 400, 300m and SWIR-15, 30m. This trend is similar to legacy protocol with the exception of the VIS-Surv domain. This is most likely because of the drastically increased template size, which improves the average quality of the faces in the templates of the VIS-Surv domain. So the CosFace model which was trained to perform well on (relatively) higher quality faces, outperforms the AdaFace model.

Please refer to the supplementary material for the 1:N open-set identification performance.
\subsection{Template Pooling - Further Analysis}
\begin{figure}[h]
	\includegraphics[width=0.5\textwidth]{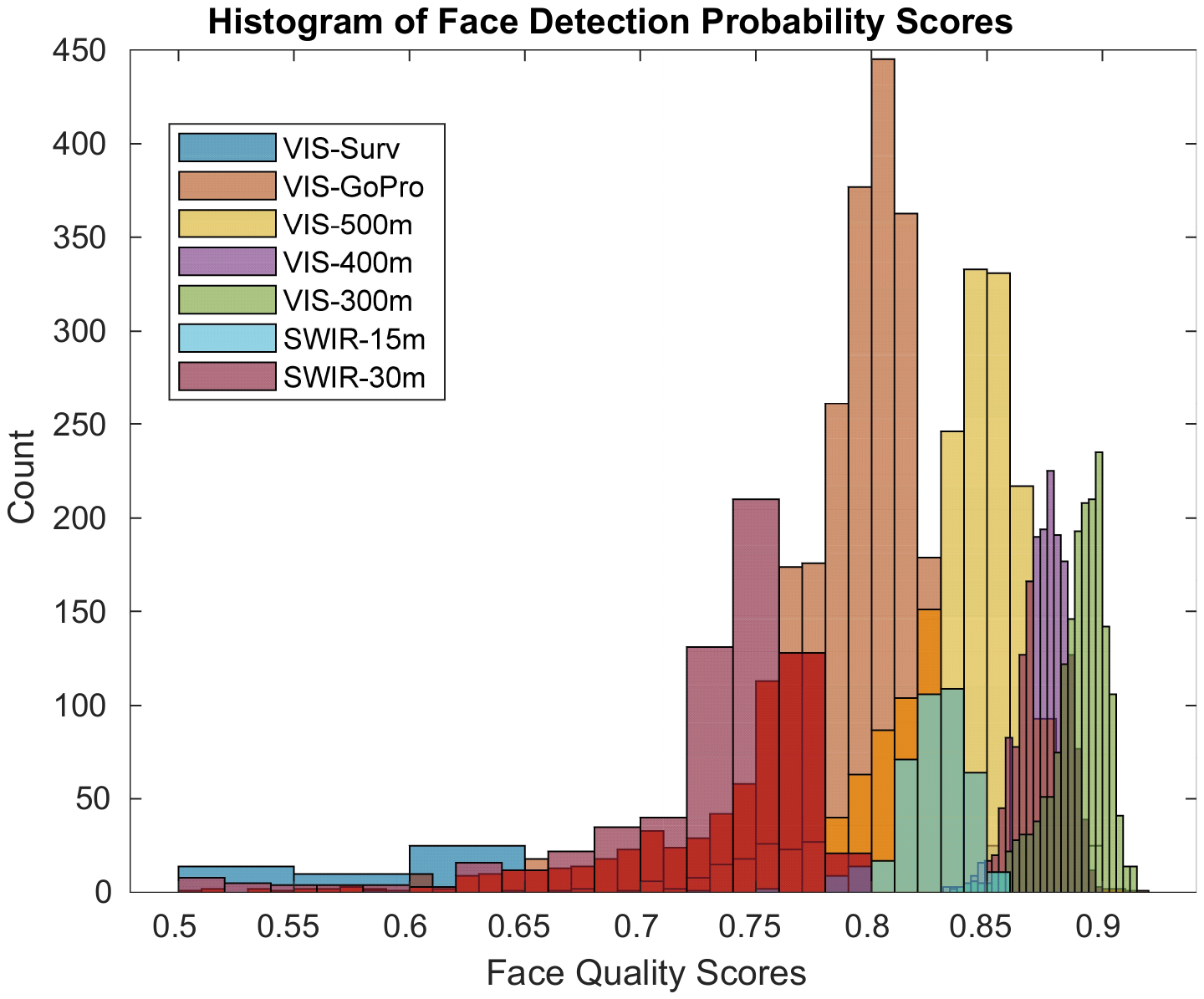}
	\caption{Histograms of face detection probability scores in a template} 
	\label{fig:hist_fq_all_doms}
	\centering
\end{figure}

\textbf{Motivation for our focus on template generation/pooling:} For the problem of cross-spectral face recognition with challenging target domains, we notice that in harder/challenging domains, there are more low quality faces and the variation in face quality is higher. We present the histograms of face detection probability scores (which denote the face quality) in a template for each of the domains in Fig. \ref{fig:hist_fq_all_doms}. We can see from the figure that the ranking of domains according to their face quality scores is as follows (decreasing order of face quality): VIS-300m, VIS-400m, VIS-500m, SWIR-15m, VIS-GoPro, SWIR-30m, and VIS-Surv. Thus, the need to properly weigh the faces in a template becomes paramount when dealing with faces from challenging domains. 

\textbf{Optimal value of $\lambda$:}
The optimal value of the hyperparameter $\lambda$ in (\ref{eq:t8}) depends both on the model and the domain of interest. Depending on the domain, the optimal value for $\lambda$ lies between 5 and 30, for the ArcFace and CosFace models. For the AdaFace model, there is a considerably higher variance - optimum $\lambda$ is between 5 and 100. Please refer to the supplementary material for more detailed results with different values of $\lambda$.

\textbf{Norm Pooling variant (NP$^*$):}
Table \ref{tab:np_np_star_w12m} presents the performance of NP$^*$ and NP on the AdaFace model (with optimal $\lambda$ values). Remarkably, NP$^*$ either equals or surpasses NP on almost all the domains except VIS-400m. Another interesting advantage of NP$^*$ is that the optimal range for $\lambda$ is between 3 and 10, a significantly smaller range compared to NP's 5-100 range. AdaFace feature norms have a relatively narrow distribution, typically between 0.8 and 1. So, to increase the spread of the media weights $c_i$, we require a larger value of $\lambda$. While max-normalization ((\ref{eq:t4})) does not alter the distribution's spread, min-max-normalization ((\ref{eq:t7})) effectively widens it by scaling the smallest norm to 0 and the largest to 1. This alleviates the burden of widening the spread from the softmax function. In contrast, the feature norms of CosFace and ArcFace models lie in a much broader range between 5 and 30. Given the already substantial spread, max-normalization proves to be sufficiently effective.

\begin{table}[t!]
	\renewcommand{\arraystretch}{1.2}
	\caption{\label{tab:np_np_star_w12m}Performance of NP$^*$ on AdaFace WebFace12M Resnet-100 Model - Rank-1 Retrieval Rates (\%)}
	\centering
	\resizebox{1\columnwidth}{!}{
		\renewcommand{\arraystretch}{1.8}
		\begin{tabular}{c c c c c c c c} 
			\hline
			\rowcolor[rgb]{0.753,0.753,0.753} \textbf{-} & \textbf{VIS-Surv} & \textbf{VIS-GoPro} & \textbf{VIS-500m} & \textbf{VIS-400m} & \textbf{VIS-300m} & \textbf{SWIR-15m} & \textbf{SWIR-30m}  \\ 
			\hline
			AP & 63.75&	59.52&	90.21&	96.46&	100.00&	94.40&	77.29 \\
			NP & 84.06 & 66.67 &  \textbf{90.91} & \textbf{97.98} & 100.00 & \textbf{95.60} & 80.08 \\
			NP$^*$ & \textbf{87.25} & \textbf{69.05} & \textbf{90.91} & 97.47 & 100.00 & \textbf{95.60} & \textbf{80.48} \\
			
			\hline
	\end{tabular}}
\end{table}

\textbf{Sparse Pooling (SP):}
We present the results of sparse pooling (SP) with the CosFace model on the exhaustive protocol in Table \ref{tab:lambda_cosface_sp}, along with the best norm pooling results. Sparse pooling outperforms or matches the performance of norm pooling on four out of the seven domains. 
The sparsity in the weights $c_i$ is directly proportional to $\lambda$. \textit{Optimal performance across all domains is achieved with average sparsity values exceeding 70\%!} This reinforces our argument for the necessity of an effective template generation algorithm, especially when working with domains having a significant amount of low-quality media.  We present further analysis of template sparsity with sparse pooling in the supplementary material.



\begin{table}[t!]
	\renewcommand{\arraystretch}{1.5}
	\caption{\label{tab:lambda_cosface_sp}Performance of SP on CosFace Glint360k Resnet-100 Model - Rank-1 Retrieval Rates (\%)}
	\centering
	\resizebox{1.\columnwidth}{!}{
		\renewcommand{\arraystretch}{1.8}
		\begin{tabular}{c c c c c c c c} 
			\hline
			\rowcolor[rgb]{0.753,0.753,0.753} \textbf{$\lambda$} & \textbf{VIS-Surv} & \textbf{VIS-GoPro} & \textbf{VIS-500m} & \textbf{VIS-400m} & \textbf{VIS-300m} & \textbf{SWIR-15m} & \textbf{SWIR-30m}  \\ 
			\hline
			AP & 74.90 & 66.67 & 81.12 & 88.89 & 99.45 & 90.00 & 66.93 \\
			NP & \textbf{98.41} & 83.33 & \textbf{86.01} & \textbf{92.93} & \textbf{100.00} & 91.60 & \textbf{72.51} \\
			SP & \textbf{98.41} & \textbf{85.71} & 84.62 & 92.42 & \textbf{100.00} & \textbf{92.80} & 71.31 \\
			
			
			\hline
	\end{tabular}}
\end{table}


\textbf{Comparison with Quality Pooling:}
To compare the performance of our proposed template pooling algorithms with quality pooling \cite{ranjan2018crystal}, we create a new protocol that uses only the faces detected by the SCR detector \cite{guo2021sample} for generating the probe templates. This is because we need the face detection probability scores to implement quality pooling. Since the detector typically detects relatively higher quality faces, we expect a much higher identification performance compared to the exhaustive protocol. The rank-1 retrieval rates (with average pooling, quality pooling, cdf pooling and norm pooling) obtained using these detections are presented in Table ~\ref{tab:e2e_results_t}. The results indicate that norm pooling (NP) outperforms quality pooling (QP) and average pooling (AP) consistently on all domains with the CosFace model.

Despite the template pooling algorithms' sensitivity to the parameter $\lambda$, and our lack of a heuristic for determining its optimal value without a validation dataset, it is noteworthy that all the proposed template pooling algorithms surpass the performance of average pooling, even with $\lambda$ set to a default value of 1. Some broad guidelines for choosing an optimal template pooling algorithm from the variants are: use min-max normalization whenever the norm distribution of a model's features is very narrow, and use sparse pooling when entirely eliminating low quality media is preferable to merely giving them smaller weights. 

A comprehensive exploration of all possible formulations of template generation algorithms using the feature-norm as a proxy for face quality is beyond the scope of this paper, but our experiments clearly show the potential for feature-norm based template pooling.

\section{Conclusion}

In this work, we introduced template generation algorithms called Norm Pooling and Sparse Pooling which can be applied even in scenarios where there is no additional training data or metadata. We showed that they outperform average pooling and quality pooling across many models and domains. We also significantly improved the 1:N face identification benchmark performance on the IJB-MDF dataset. Our results indicate that the efficacy of models (CosFace, ArcFace or AdaFace) is domain dependent. Notably, models trained with AdaFace loss perform better on long-range/remote domains (faces captured at a distance), while models trained with ArcFace and CosFace losses surpass the AdaFace models on surveillance and go-pro domains. \par


\par
\section*{Acknowledgement}
\small{This research is based upon work supported by the Office
of the Director of National Intelligence (ODNI), Intelligence
Advanced Research Projects Activity (IARPA), via IARPA R\&D
Contract No. 2019-022600002. The views and conclusions contained
herein are those of the authors and should not be interpreted
as necessarily representing the official policies or endorsements,
either expressed or implied, of the ODNI, IARPA,
or the U.S. Government. The U.S. Government is authorized to
reproduce and distribute reprints for Governmental purposes
notwithstanding any copyright annotation thereon.}

\clearpage
{\small
\bibliographystyle{ieee}
\bibliography{egbib}
}
\end{document}